\documentclass[sigconf]{acmart}

\copyrightyear{2026}
\acmYear{2026}
\setcopyright{cc}
\setcctype{by}
\acmConference[WWW Companion '26]{Companion Proceedings of the ACM Web Conference 2026}{April 13--17, 2026}{Dubai, United Arab Emirates}
\acmBooktitle{Companion Proceedings of the ACM Web Conference 2026 (WWW Companion '26), April 13--17, 2026, Dubai, United Arab Emirates}
\acmPrice{}
\acmDOI{10.1145/3774905.3795609}
\acmISBN{979-8-4007-2308-7/2026/04}

\usepackage{graphicx}
\usepackage{booktabs}
\usepackage{tabularx}
\usepackage{array}
\usepackage{amsmath}
\usepackage{xcolor}
\usepackage{enumitem}
\usepackage{listings}
\usepackage{float}
\usepackage{placeins}

\graphicspath{{./}{./figures/}{./images/}}

\setlist[itemize]{leftmargin=*, itemsep=2pt, topsep=2pt}
\setlist[enumerate]{leftmargin=*, itemsep=2pt, topsep=2pt}

\title[Infographic Reconstruction to Slides]{From Dead Pixels to Editable Slides: Infographic Reconstruction into Native Google Slides via Vision-Language Region Understanding}

\author{Leonardo Gonzalez}
\affiliation{%
  \institution{Trilogy AI Center of Excellence}
  \city{Austin}
  \country{USA}
}

\begin{abstract}
Infographics are widely used to communicate information with a combination of text, icons, and data visualizations, but once exported as images their content is locked into pixels, making updates, localization, and reuse expensive.
We describe \textsc{Images2Slides}, an API-based pipeline that converts a static infographic (PNG/JPG) into a native, editable Google Slides slide by extracting a region-level specification with a vision-language model (VLM), mapping pixel geometry into slide coordinates, and recreating elements using the Google Slides batch update API.
The system is model-agnostic and supports multiple VLM backends via a common JSON region schema and deterministic postprocessing.
On a controlled benchmark of 29 programmatically generated infographic slides with known ground-truth regions, \textsc{Images2Slides} achieves an overall element recovery rate of $0.989\pm0.057$ (text: $0.985\pm0.083$, images: $1.000\pm0.000$), with mean text transcription error $\mathrm{CER}=0.033\pm0.149$ and mean layout fidelity $\mathrm{IoU}=0.364\pm0.161$ for text regions and $0.644\pm0.131$ for image regions.
We also highlight practical engineering challenges in reconstruction, including text size calibration and non-uniform backgrounds, and describe failure modes that guide future work.
\end{abstract}



\begin{document}
\maketitle

\section{Introduction}
Infographics compress complex information into compact visuals that mix dense text, graphical annotations, and small icons or charts.
Once exported as raster images, however, they become \emph{dead pixels}: updating labels, correcting numbers, translating text, or repurposing the design for new audiences typically requires redesign work.

Recent progress in vision-language models (VLMs) suggests that screen-like media can be parsed into structured, element-level representations that preserve content and geometry.
ScreenAI, for example, targets UIs and infographics and is trained with a screen annotation task that predicts element types and locations \cite{screenai}.
Benchmarks such as DocVQA and InfographicVQA emphasize the need to jointly reason over embedded text and layout \cite{mathew2021docvqa,mathew2022infographicvqa}, while newer suites like MMMU and MMDocBench broaden evaluation across diverse multimodal inputs, including charts and infographics \cite{yue2024mmmu,mmdocbench}.
Separately, ``derendering'' has emerged as a useful lens for converting visual artifacts into structured, editable representations \cite{lee2023pix2struct,si2024design2code}.

This paper presents \textsc{Images2Slides}, an implemented pipeline that reconstructs a static infographic as a native Google Slides slide using explicit regions and API-driven creation \cite{slidesapi}.
The core goal is editability: textual content becomes editable slide text objects while preserving layout as closely as possible.

\textbf{Contributions.} We contribute:
\begin{itemize}
  \item An end-to-end, API-based system that reconstructs an infographic image as an editable Google Slides slide via region extraction, geometry mapping, and structured slide generation \cite{slidesapi}.
  \item A lightweight, model-agnostic region schema with deterministic validation and postprocessing, enabling multiple VLM backends (including open-weight alternatives such as Qwen3-VL) \cite{qwen3vl}.
  \item Practical reconstruction techniques, including deterministic element IDs for retries and a piecewise-linear font size calibration that improves readability of small text.
  \item A discussion of limitations and workflow strategies for non-uniform backgrounds and raster-only visual elements, informed by related work in document layout analysis and chart understanding \cite{dlaformer,zhong2019publaynet,masry2022chartqa,liu2023deplot,masry2023unichart}.

  \item A quantitative evaluation on a controlled benchmark with programmatically generated ground truth, reporting text accuracy (CER/WER), layout fidelity (IoU and center offset), and editability recovery rates.
\end{itemize}

\section{Related Work}
Our approach relates to three research threads: (i) multimodal understanding of documents and infographics; (ii) region extraction and layout reconstruction; and (iii) visual-to-structured generation and derendering.

\paragraph{Multimodal document and infographic understanding.}
Document VQA benchmarks highlight the interaction between layout, text, and visuals \cite{mathew2021docvqa,mathew2022infographicvqa}.
General multimodal benchmarks like MMMU test broad perception and reasoning, including diagram and chart-style inputs \cite{yue2024mmmu}.
MMDocBench provides OCR-free document tasks with supporting regions, enabling fine-grained evaluation of localization and extraction in charts and infographics \cite{mmdocbench}.
Model-side work addresses the difficulty of reading text-rich images: DocOwl 1.5 emphasizes structure learning for OCR-free document understanding \cite{hu2024docowl15}, while DocVLM augments frozen VLMs with OCR-derived text and layout as learned queries to reduce reliance on high-resolution visual tokens \cite{docvlm}.
ScreenAI focuses specifically on UIs and infographics and uses element-level supervision for types and locations \cite{screenai}.

\paragraph{Region extraction and layout reconstruction.}
Document layout analysis decomposes pages into regions and reading order, providing a basis for conversion and retrieval.
Large datasets such as PubLayNet and DocLayNet enable scalable training and evaluation of layout detectors \cite{zhong2019publaynet,pfitzmann2022doclaynet}.
Tooling such as LayoutParser streamlines application of deep layout models \cite{shen2021layoutparser}.
End-to-end approaches like DLAFormer integrate multiple layout subtasks in a transformer architecture by casting detection, classification, and ordering as relation prediction \cite{dlaformer}.
Complementary work on multimodal document pretraining, such as LayoutLM/LayoutLMv3 and UDOP, unifies text, layout, and vision signals for downstream document tasks \cite{xu2020layoutlm,huang2022layoutlmv3,tang2023udop}.

\paragraph{Visual-to-structured generation and derendering.}
OCR-free sequence generation approaches such as Donut output structured representations directly from document images \cite{kim2022donut}.
Nougat focuses on converting scientific PDFs into structured markup, recovering layout-dependent semantics like math \cite{blecher2024nougat}.
Pix2Struct uses screenshot parsing as pretraining, learning to map masked screenshots to simplified HTML \cite{lee2023pix2struct}.
In chart-specific settings, ChartQA benchmarks question answering over charts \cite{masry2022chartqa}, and models such as MatCha, DePlot, UniChart, and ChartReader study chart derendering and chart-to-structured representations \cite{liu2023matcha,liu2023deplot,masry2023unichart,cheng2023chartreader}.
Design2Code benchmarks multimodal generation of front-end code from visual designs \cite{si2024design2code}.
We adopt a related perspective but target the Google Slides object model as the structured output.

\section{Problem Formulation}
Given an infographic image $I$ with pixel dimensions $(W_I, H_I)$, we aim to produce a Google Slides slide $S$ containing a set of page elements $\{e_k\}$ such that:
\begin{enumerate}
  \item \textbf{Editability:} text in $I$ is recreated as editable slide text objects.
  \item \textbf{Layout fidelity:} the spatial arrangement of elements approximates the original infographic geometry.
  \item \textbf{Practicality:} the pipeline is robust to varied infographic styles and supports iterative refinement.
\end{enumerate}

We represent infographic structure as a region file $\mathcal{R}$, produced by a VLM and validated by deterministic postprocessing.
Each region includes an ID; a type (\texttt{text} or \texttt{image}); a pixel bounding box; optional reading order and tags; extracted text for \texttt{text} regions; and optional style hints and confidence.

\section{System Overview}
\begin{figure}[t]
  \centering
  \includegraphics[width=0.95\linewidth,trim=10 10 10 0,clip]{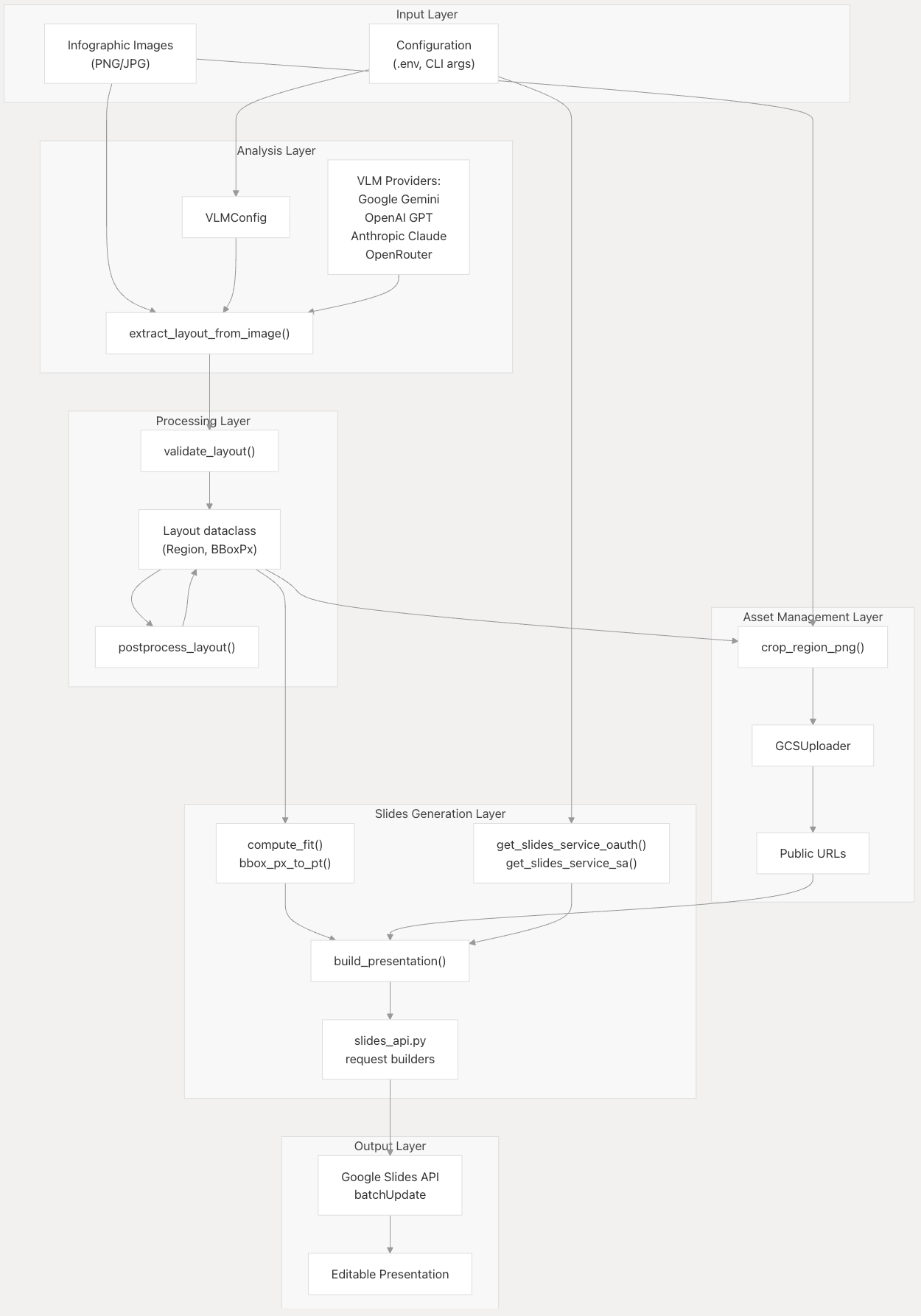}
  \caption{Layered architecture of \textsc{Images2Slides}. The pipeline is organized into input, analysis, processing, asset management, slides generation, and output layers.}
  \Description{A layered block diagram showing how infographic images and configuration flow through VLM analysis, validation and postprocessing, asset preparation, geometry mapping, request generation, and Google Slides API execution to produce an editable presentation.}
  \label{fig:layers}
\end{figure}

Figure~\ref{fig:layers} summarizes the modular architecture of \textsc{Images2Slides}. Table~\ref{tab:dataflow} summarizes the data flow.

\begin{table*}[t]
\centering
\caption{Data flow through the pipeline (stages, artifacts, and outputs).}
\label{tab:dataflow}
\footnotesize
\setlength{\tabcolsep}{5pt}
\renewcommand{\arraystretch}{1.15}
\begin{tabularx}{\textwidth}{@{}p{0.12\textwidth}p{0.18\textwidth}X p{0.20\textwidth}@{}}
\toprule
\textbf{Stage} & \textbf{Input} & \textbf{Processing} & \textbf{Output} \\
\midrule
1. VLM analysis &
Infographic image (PNG/JPG) &
\texttt{extract\_layout\_from\_image()} using a selected VLM backend and a strict JSON schema (regions, types, bboxes, text). &
Raw JSON layout (text and image regions) \\
\addlinespace[2pt]
2. Validation \& postprocess &
Raw JSON layout &
\texttt{validate\_layout()} converts JSON to typed objects; \texttt{postprocess\_layout()} normalizes text, clamps boxes, applies minimal heuristics for consistency. &
Clean, typed \texttt{Layout} object \\
\addlinespace[2pt]
3. Asset prep &
Image + \texttt{Layout} &
For each image region: \texttt{crop\_region\_png()} (with small padding) and upload to obtain an HTTPS-accessible URL. &
Public URLs for image regions \\
\addlinespace[2pt]
4. Geometry &
\texttt{Layout} + slide page size &
\texttt{compute\_fit()} preserves aspect ratio and centers the image; \texttt{bbox\_px\_to\_pt()} maps pixel boxes to slide points. &
Slide coordinates (pt) for all regions \\
\addlinespace[2pt]
5. API requests &
Coordinates + asset URLs &
\texttt{build\_requests\_for\_infographic()} constructs a single Slides API batch (create slide, add text boxes, insert text, place images). &
Slides API request batch \\
\addlinespace[2pt]
6. Execution &
Request batch &
\texttt{presentations.batchUpdate} applies the batch to the target presentation. &
Live, editable Google Slides presentation \\
\bottomrule
\end{tabularx}
\end{table*}

\subsection{VLM-based region extraction}
We prompt a VLM to produce a strict JSON region file $\mathcal{R}$ with pixel-space coordinates, region types, and extracted text.
The prompt emphasizes explicit image dimensions, well-scoped regions, stable IDs, and reading order and tags when available.
This aligns with element-centric screen understanding \cite{screenai} and region-grounded evaluation in document benchmarks \cite{mmdocbench}.

\subsection{Region JSON schema}
\label{sec:schema}
To support reliable reconstruction across model backends, \textsc{Images2Slides} uses a strict, typed region schema that distinguishes two region types: \texttt{text} and \texttt{image}.
Both include pixel-space geometry, while \texttt{text} regions also carry extracted text and optional font/style hints, and \texttt{image} regions specify only the coordinates used to crop the source image.
The top-level JSON contains image dimensions and a list of regions:
\begin{lstlisting}[caption={Top-level layout schema (abridged).},label={lst:schema}]
{
  "image_px": {
    "width": 1600,
    "height": 900
  },
  "regions": [
    {
      "id": "title",
      "order": 1,
      "type": "text",
      "bbox_px": {
        "x": 35.0,
        "y": 45.0,
        "w": 1500.0,
        "h": 60.0
      },
      "text": "The Psychology of Buying: 3 Triggers Driving E-Commerce Sales",
      "style": {
        "font_family": "Arial",
        "font_size_pt": 42,
        "bold": true
      },
      "crop_from_infographic": false,
      "confidence": 0.99,
      "notes": null
    },
    {
      "id": "image_social_proof",
      "order": 2,
      "type": "image",
      "bbox_px": {
        "x": 32.0,
        "y": 140.0,
        "w": 469.0,
        "h": 351.0
      },
      "text": null,
      "style": null,
      "crop_from_infographic": true,
      "confidence": 0.98,
      "notes": "Includes smartphone illustration and background container"
    }
  ]
}
\end{lstlisting}

Table~\ref{tab:region_schema} summarizes the region object fields and key constraints enforced during validation and postprocessing.

\begin{table}[t]
\centering
\caption{Region schema (abridged) and validation rules.}
\label{tab:region_schema}
\footnotesize
\setlength{\tabcolsep}{4pt}
\renewcommand{\arraystretch}{1.15}
\begin{tabularx}{\linewidth}{@{}l X@{}}
\toprule
\textbf{Field} & \textbf{Description / constraints} \\
\midrule
\texttt{id} & Required. Stable identifier (string) used for deterministic object IDs in Slides (e.g., \texttt{TXT\_<id>}). \\
\texttt{order} & Required. Reading order hint (integer); fallback ordering is computed if missing. \\
\texttt{type} & Required. One of \{\texttt{text}, \texttt{image}\}. \\
\texttt{bbox\_px} & Required. Pixel-space box \texttt{\{x,y,w,h\}} in the input image coordinate system; clamped to image bounds; requires $w,h>0$. \\
\texttt{text} & Required for \texttt{type=text}. Must be non-empty after whitespace normalization. \\
\texttt{style} & Optional style hints (e.g., \texttt{font\_family}, \texttt{font\_size\_pt}, \texttt{bold}); missing fields fall back to defaults. \\
\texttt{crop\_from\_infographic} & Optional boolean. If true, the pipeline crops the region from the original infographic and inserts it as an image object. \\
\texttt{confidence}, \texttt{notes} & Optional. Confidence in $[0,1]$ and free-form notes for debugging. \\
\bottomrule
\end{tabularx}
\end{table}

\noindent We constrain prompting to improve parseability: the model must output \emph{JSON only} (no Markdown or commentary), include exact image dimensions, keep all boxes within bounds, and avoid inferring or fabricating text that is not visible in the image.

The implementation supports multiple VLM providers behind a unified interface, including Google Gemini \cite{gemini3pro}, OpenAI GPT, Anthropic Claude, and OpenRouter-hosted models (e.g., Qwen3-VL) \cite{qwen3vl}.
To improve robustness, malformed or schema-violating JSON responses are detected during validation and can trigger a retry with error feedback describing the expected schema.

\subsection{Layout postprocessing}
Model outputs can contain minor inconsistencies (e.g., empty strings, boxes outside bounds).
We apply deterministic postprocessing: whitespace normalization, bounds clamping, minimum size enforcement, and fallback ordering (top-to-bottom, left-to-right).
Intermediate artifacts (raw JSON, validated JSON, rendered debug overlays) are cached to support rapid iteration.

\subsection{Geometry mapping}
Slides use page coordinates in points, while regions are defined in pixels.
Object geometry is expressed via size and affine transforms in the Slides API \cite{slidestransforms}.
We fit the infographic into the slide while preserving aspect ratio and compute a scale factor and centering offsets.
Let slide page size be $(W_S, H_S)$ in points, and image size be $(W_I, H_I)$ in pixels:
\begin{align}
  s &= \min\left(\frac{W_S}{W_I}, \frac{H_S}{H_I}\right), \\
  \Delta_x &= \frac{W_S - sW_I}{2}, \quad
  \Delta_y = \frac{H_S - sH_I}{2}.
\end{align}
A region with pixel box $(x, y, w, h)$ maps to slide rectangle

$(x', y', w', h') = (\Delta_x + sx,\ \Delta_y + sy,\ sw,\ sh)$.

\subsection{Slide reconstruction via Google Slides API}
We create and populate slides using \texttt{presentations.batchUpdate} \cite{slidesapi}.
The pipeline issues a batch of API requests to create a slide, add text boxes, insert text, and apply text styling.
For image regions, we crop the region from the original infographic, upload it to an HTTPS-accessible URL, and place it as an image element.

\paragraph{Deterministic IDs for retries.}
To support retries and incremental updates, the system uses deterministic object IDs derived from region IDs (e.g., \texttt{TXT\_<id>}, \texttt{IMG\_<id>}).

\section{Typography Calibration}
\label{sec:typography}
In practice, VLM-based region extraction tends to underestimate font sizes, especially for small body text and labels.
Because slide geometry is scaled from image pixels into slide points, the system first derives a base font size in points,
$f = f_{\text{vlm}} \cdot s$, where $f_{\text{vlm}}$ is the model-provided estimate and $s$ is the image-to-slide scale factor used for placement.

\subsection{Piecewise-linear font scaling}
We apply a piecewise-linear calibration that boosts smaller fonts while leaving large titles unchanged.
Let $f$ be the base font size (in points). The calibrated size is:
\begin{equation}
  f' = f + \max\!\bigl(0,\ (14 - f)\times 0.294\bigr).
  \label{eq:font_scaling}
\end{equation}
This function has two anchor behaviors that match the implemented workflow:
(i) small body text at 5.5\,pt is increased to 8\,pt (a readable minimum in many slide contexts), and
(ii) text at 14\,pt and above is left unchanged.
Below 14\,pt, the mapping is linear and provides larger relative boosts to smaller fonts.

\subsection{Collision-aware text box width expansion}
Increasing font size can increase the rendered line length and cause overflow within a fixed-width text box.
To mitigate this, we optionally expand the text box width in proportion to the font scaling factor while preventing overlap with neighboring regions:
\begin{enumerate}
  \item Compute the font ratio $r = f'/f$.
  \item Compute a desired width $w'_{\text{desired}} = w' \cdot r$ (where $w'$ is the mapped box width in points).
  \item Initialize the candidate right edge $\textit{right} = x' + w'_{\text{desired}}$.
  \item Cap $\textit{right}$ to the slide boundary with a small margin.
  \item If there is a neighboring region to the right whose vertical span overlaps this text box, cap $\textit{right}$ to that region's left edge minus a small gap.
  \item Set the adjusted width $w'_{\text{adj}} = \max(w',\ \textit{right}-x')$ while keeping the left edge $x'$ fixed.
\end{enumerate}
This preserves alignment while reducing overflow and avoiding collisions in dense layouts.
\section{Image Region Handling}
Non-text regions (icons, logos, small diagrams, and chart insets) are handled as raster assets: we crop the region from the source infographic and place it as an image element in Slides.

\subsection{Crop padding}
VLMs often produce tight bounding boxes that can clip strokes or antialiased edges.
We therefore pad the crop on the right and bottom edges by a small fixed amount (10\,px in the current implementation), clamped to image bounds:
\begin{align}
  x_2 &= \min(x + w + 10,\ W_I), \\
  y_2 &= \min(y + h + 10,\ H_I).
\end{align}
The top-left corner $(x,y)$ is preserved to maintain alignment with overlaid elements.

\subsection{Upload and deduplication}
The Google Slides API creates images from URLs, so cropped assets must be uploaded to an HTTPS-accessible location before slide creation \cite{slidesapi}.
To reduce redundant uploads and improve reproducibility, cropped images are stored with content-addressed names (a hash of the image bytes), enabling deduplication and caching across runs.

\section{Background Handling}
\label{sec:bg}
Non-uniform or textured backgrounds complicate reconstruction, particularly when the goal is a clean, editable slide.
\textsc{Images2Slides} optionally synthesizes a clean background from the infographic when the \texttt{--synthesize-background} flag is enabled.
The pipeline prompts a VLM to return a \texttt{background\_sample} JSON object with a pixel-space box and a mode label:
\texttt{solid} for near-uniform backgrounds or \texttt{tile} for textured/patterned ones.
We crop the sampled patch and generate a full-size background image by either (i) filling the slide with the patch's average color (\texttt{solid}) or
(ii) tiling the patch across the slide (\texttt{tile}).
The synthesized background is uploaded and inserted as a single image element, then editable text and cropped region images are placed above it.

When background synthesis is disabled, we omit the background entirely and reconstruct only extracted text and image regions to avoid duplicating foreground content.

\FloatBarrier
\begin{table*}[!t]
\centering
\caption{Quantitative evaluation on 29 programmatically generated infographic slides. Values are mean $\pm$ std across runs.}
\label{tab:eval_results}
\footnotesize
\setlength{\tabcolsep}{6pt}
\renewcommand{\arraystretch}{1.15}
\begin{tabular}{lccc}
\toprule
\textbf{Metric} & \textbf{Text} & \textbf{Image} & \textbf{Overall / global} \\
\midrule
Element recovery rate & $0.985\pm0.083$ & $1.000\pm0.000$ & $0.989\pm0.057$ \\
Character recovery rate & $0.969\pm0.124$ & -- & -- \\
Mean IoU & $0.364\pm0.161$ & $0.644\pm0.131$ & -- \\
Median IoU & $0.388\pm0.177$ & $0.639\pm0.130$ & -- \\
Mean center offset (px) & $53.7\pm26.9$ & $11.0\pm16.3$ & -- \\
Mean CER / WER & $0.033\pm0.149$ / $0.037\pm0.167$ & -- & -- \\
Frac.\ IoU $\ge 0.5$ & $0.376\pm0.292$ & $0.871\pm0.332$ & $0.513\pm0.237$ \\
Frac.\ IoU $\ge 0.75$ & $0.031\pm0.092$ & $0.181\pm0.362$ & $0.071\pm0.110$ \\
VLM extraction time (s) & \multicolumn{3}{c}{$54.999\pm18.337$} \\
Slides API time (s) & \multicolumn{3}{c}{$5.792\pm0.682$} \\
\bottomrule
\end{tabular}
\end{table*}
\FloatBarrier
\section{Evaluation}
\label{sec:eval}
A common challenge in evaluating raster-to-editable pipelines is obtaining ground-truth element geometry.
To avoid manual labeling, we use a controlled round-trip benchmark that programmatically generates ground-truth Slides with known region coordinates, rasterizes them, and then reconstructs them with \textsc{Images2Slides}.

\subsection{Benchmark construction}
\label{sec:eval_benchmark}
Each evaluation run generates a synthetic infographic slide by (i) sampling a grid/panel layout template, (ii) prompting a text VLM to produce a cohesive infographic concept and a region JSON specification conforming to our schema (Section~\ref{sec:schema}), and (iii) generating the referenced images with a text-to-image model.
The ground-truth slide is created using the same Slides generation functions as \textsc{Images2Slides}, ensuring that region boxes correspond to known pixel coordinates in the exported PNG.
We then export the slide as a raster image and feed it to \textsc{Images2Slides} to obtain a predicted region JSON and an editable reconstructed slide.

We report results over 29 successfully completed runs (provider: Google Gemini backend), with each run producing a single $1600\times 900$ infographic image and paired ground truth and prediction \texttt{gt\_region.json} and \texttt{pred\_region.json} files.

We report quantitative results for a single backend to provide a concrete reference point; the benchmarking harness is backend-agnostic and can be rerun with alternative providers.
We include wall-clock time for the VLM and Slides API stages.
Future work could log input/output token counts from provider response metadata, where available, to enable post-hoc cost estimation.

\subsection{Metrics}
\label{sec:eval_metrics}
We evaluate three aspects of reconstruction:

\begin{itemize}
  \item \textbf{Editability recovery.} Element recovery rate is the fraction of ground-truth regions that are matched by a predicted region of the same type.
  We also compute character recovery rate, the fraction of ground-truth characters recovered across matched text regions.
  \item \textbf{Layout fidelity.} For each matched region we compute bounding-box Intersection-over-Union (IoU) in pixel space and center-point displacement (in pixels and normalized by image diagonal).
  \item \textbf{Text accuracy.} For matched text regions we compute character error rate (CER) and word error rate (WER) via Levenshtein distance.
\end{itemize}

\noindent Predicted regions are matched to ground-truth regions using a one-to-one assignment within each region type that maximizes overlap (IoU); unmatched predictions are counted as false positives, and unmatched ground-truth regions as false negatives.
\vspace{-8pt}
\subsection{Results}
Table~\ref{tab:eval_results} summarizes mean and standard deviation across runs.
Overall element recovery is high ($0.989\pm0.057$), with perfect image-region recovery and occasional text-region over-segmentation (false positives) or misses (false negatives).
Layout fidelity is stronger for image regions than for text regions, consistent with the fact that small changes in font metrics, line breaking, and text box width can produce low IoU even when transcription is correct.

Across all matched elements (globally aggregated), $37.6\%$ of text boxes and $85.1\%$ of image boxes achieve IoU $\ge 0.5$, while $2.8\%$ of text boxes and $15.8\%$ of image boxes achieve IoU $\ge 0.75$.
Mean transcription error is low (CER $0.033\pm0.149$, WER $0.037\pm0.167$), but variance indicates that a minority of runs still incur noticeable recognition errors (typically stylized or low-contrast text).
\vspace{-8pt}
\subsection{Failure modes}
\label{sec:failure_modes}
We observed recurring failure patterns that explain most residual errors:

\begin{table}[H]
\centering
\caption{Observed failure modes and mitigations.}
\label{tab:failures}
\footnotesize
\setlength{\tabcolsep}{3pt}
\renewcommand{\arraystretch}{1.0}
\begin{tabularx}{\linewidth}{@{}l X@{}}
\toprule
\textbf{Failure mode} & \textbf{Symptom and mitigation} \\
\midrule
Text over-segmentation & A single paragraph is split into multiple regions, increasing false positives and lowering IoU. Mitigation: merge adjacent text boxes with compatible styles and small gaps. \\
Text under-segmentation & Multiple logical blocks are merged, harming editability. Mitigation: split using OCR line clustering or whitespace-based heuristics. \\
Font metric mismatch & Correct text but incorrect line breaks and box sizes reduce IoU. Mitigation: font calibration (Section~\ref{sec:typography}) and collision-aware width expansion. \\
Low contrast / stylized text & Non-zero CER/WER in runs with decorative typography or low resolution. Mitigation: upscale preprocessing, contrast enhancement, or fallback to OCR for text regions. \\
Non-uniform backgrounds & Decorative containers or gradients complicate clean reconstruction. Mitigation: VLM-guided background sampling and synthesized background tiling/fill (Section~\ref{sec:bg}). \\
\bottomrule
\end{tabularx}
\end{table}

\clearpage
\subsection{Qualitative example}
\begin{figure}[t]
  \centering
  \includegraphics[width=0.95\linewidth]{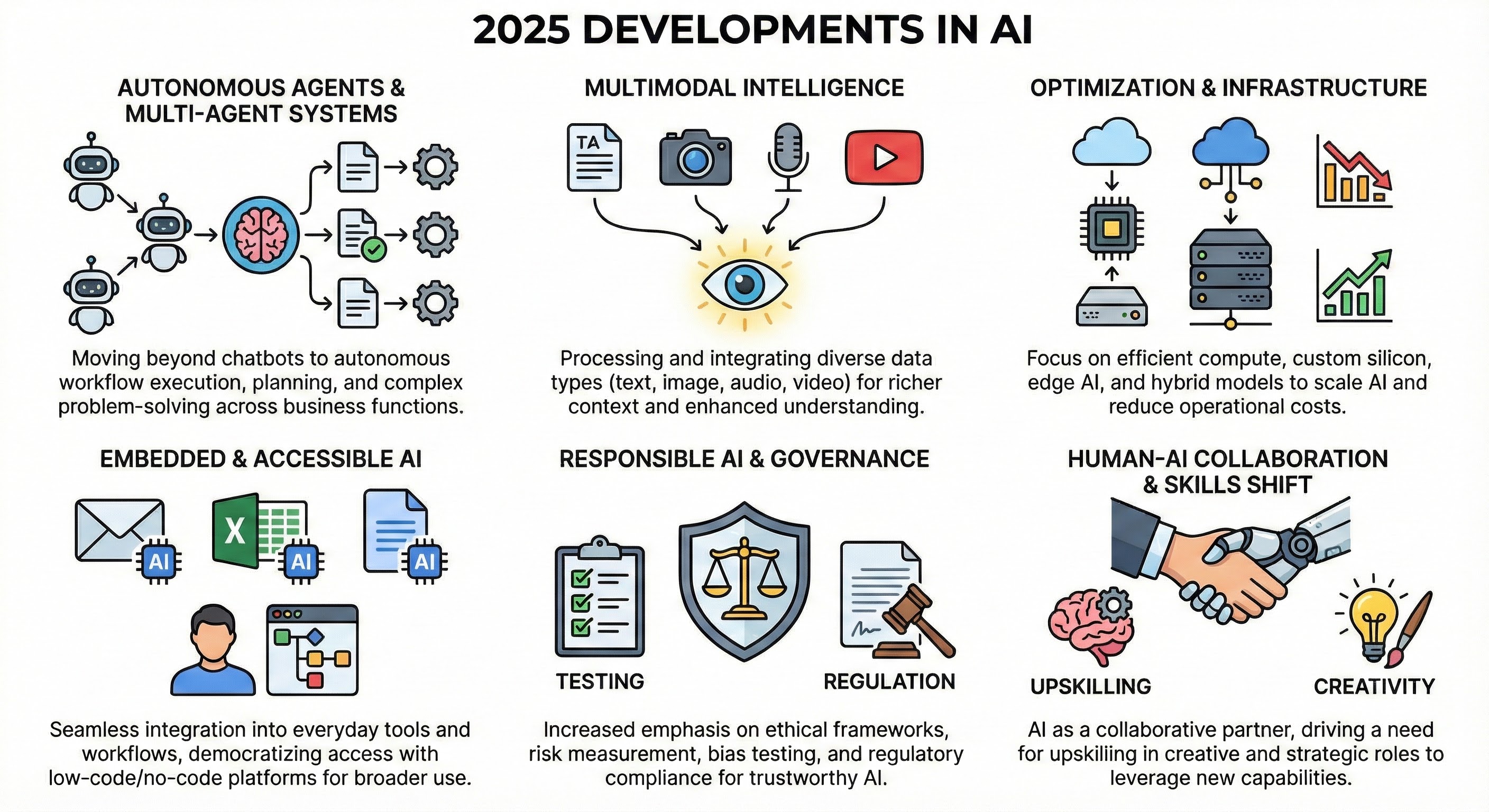}\\[3pt]
  {\footnotesize (a) Input infographic.}\\[6pt]
  \includegraphics[width=0.95\linewidth]{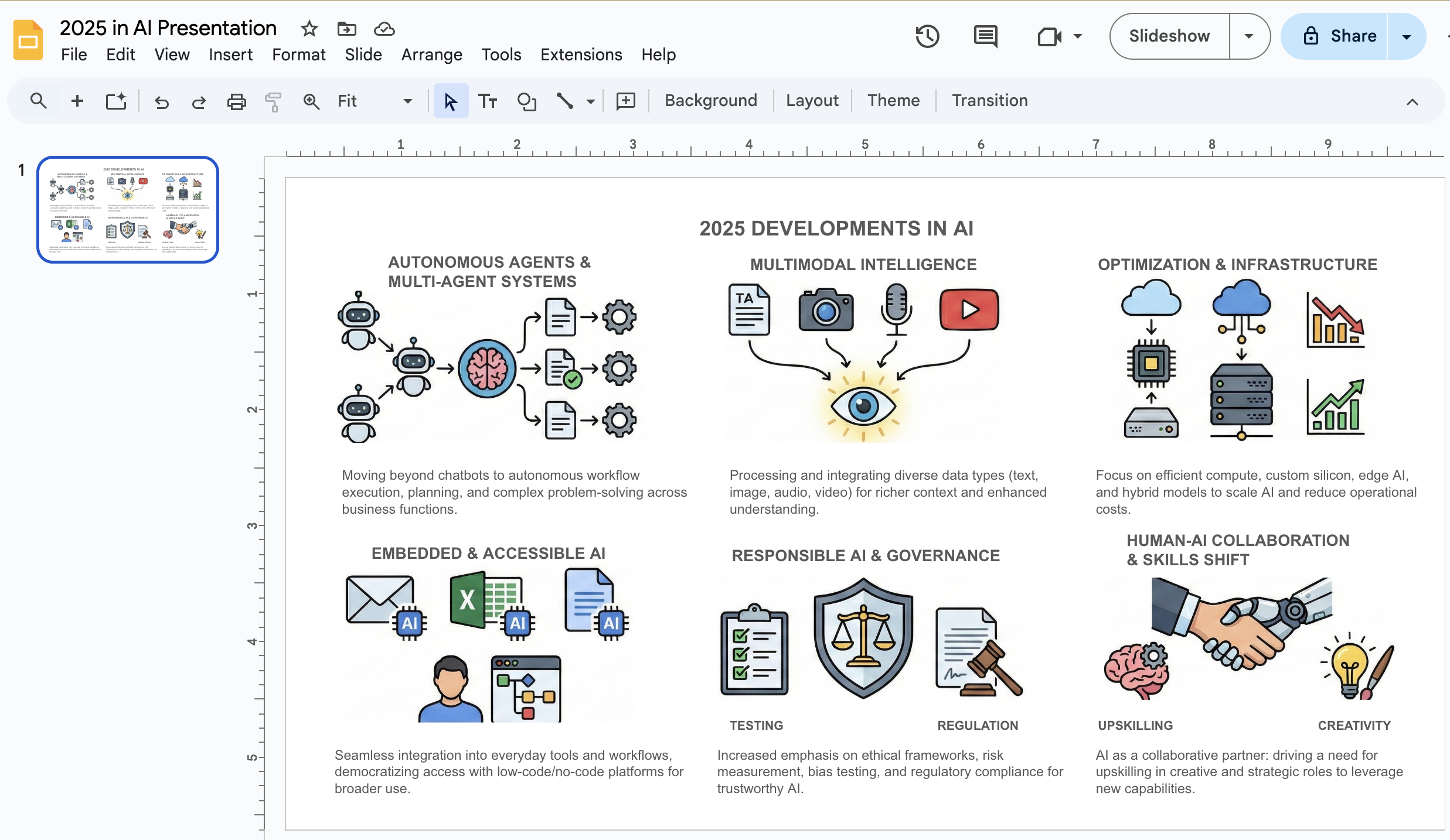}\\[3pt]
  {\footnotesize (b) Reconstructed Google Slide.}\\[6pt]
  \includegraphics[width=0.95\linewidth]{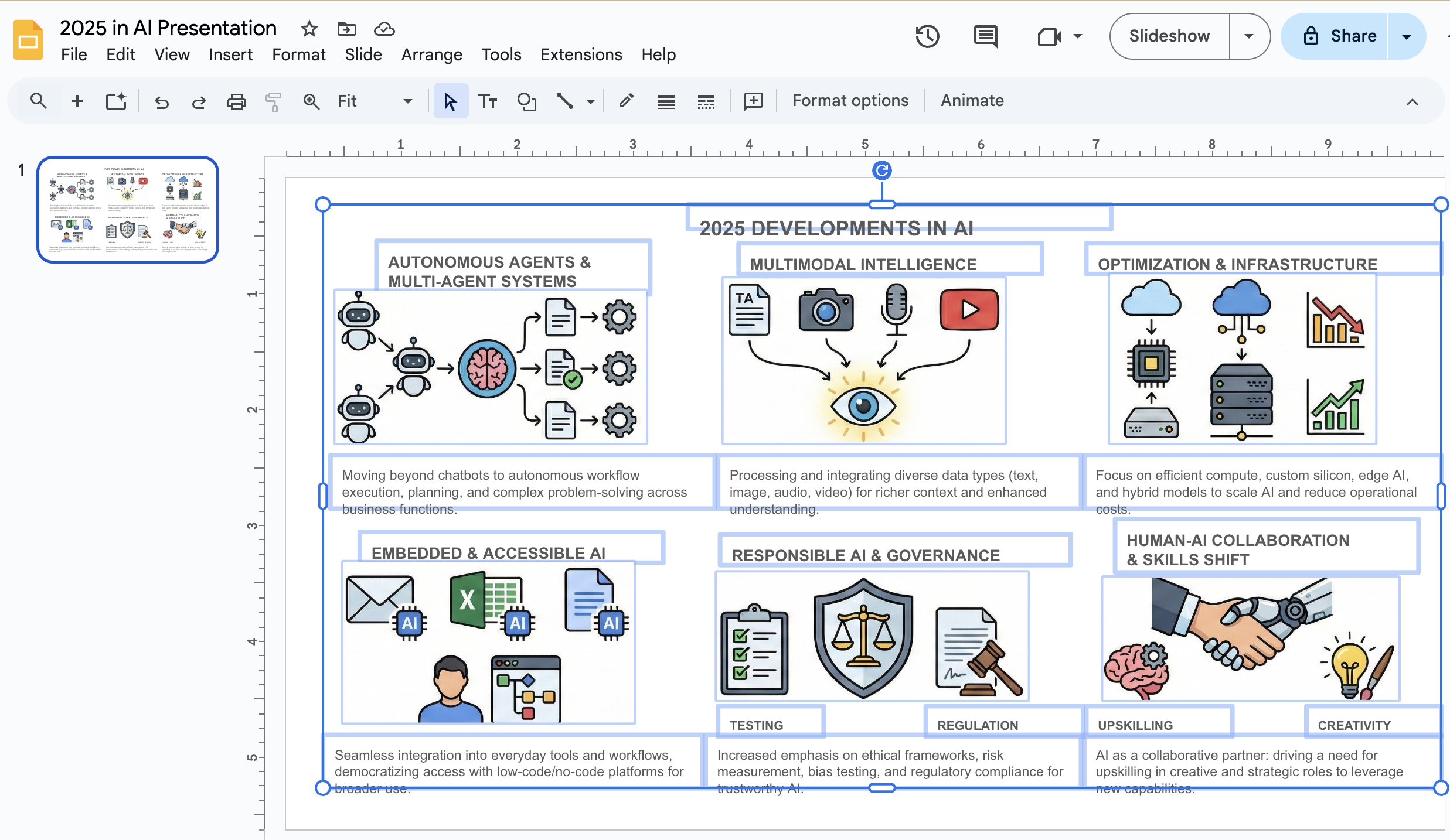}\\[3pt]
  {\footnotesize (c) Slide with regions selected.}
  \caption{End-to-end example reconstruction produced by \textsc{Images2Slides}.}
  \Description{Stacked images showing an input infographic, the reconstructed Google Slide, and the reconstructed slide with regions selected to visualize recovered element geometry.}
  \label{fig:demo_stack}
\end{figure}

Figure~\ref{fig:demo_stack} illustrates a representative end-to-end reconstruction: the input infographic, the reconstructed Google Slides slide, and the same slide with all reconstructed regions selected to visualize element boundaries.

\section{Discussion and Limitations}
\paragraph{Raster-to-native gap.}
The pipeline prioritizes editability of text and placement of raster visual elements.
It does not fully reconstruct complex vector primitives (custom shapes, gradients, rich charts) as native slide objects.
Chart-specific derendering methods \cite{liu2023matcha,liu2023deplot,masry2023unichart,cheng2023chartreader} suggest a path toward future chart-to-native regeneration.

\paragraph{Dependence on region quality.}
Reconstruction quality depends on the region specification.
Layout and structure learning methods \cite{dlaformer,hu2024docowl15,docvlm} motivate improvements in localization fidelity and ordering that could directly benefit reconstruction.

\paragraph{Operational constraints.}
The Slides API requires publicly accessible URLs for images, which introduces an asset-hosting component in end-to-end automation \cite{slidesapi}.

\section{Ethical Considerations}
This work can be applied to copyrighted or sensitive visual content; deployments should respect content provenance, licensing, and privacy.
When used with generated infographics, systems should disclose synthetic origins and avoid deceptive presentation.

\section{Artifact Availability}
Code, evaluation scripts, and assets are available at \url{https://github.com/kumanday/images2slides}.

\section{Conclusion}
We presented \textsc{Images2Slides}, an implemented pipeline for converting static infographic images into native, editable Google Slides slides via VLM-based region extraction and API-driven reconstruction.
The approach bridges document-style visual understanding and practical visualization authoring, emphasizing editability and workflow usability.
We documented engineering techniques that improve reconstruction in practice, particularly for typography and backgrounds.

\bibliographystyle{ACM-Reference-Format}

\end{document}